\documentclass{article}
\usepackage[final]{neurips_2021}

\usepackage{times}
\usepackage{latexsym}

\usepackage[utf8]{inputenc} 
\usepackage[T1]{fontenc}    
\usepackage{hyperref}       
\usepackage{url}            
\usepackage{booktabs}       
\usepackage{amsfonts}       
\usepackage{amsmath}
\usepackage{nicefrac}       
\usepackage{microtype}      
\usepackage{xcolor}         
\usepackage{subfig}
\usepackage{graphicx}
\usepackage{multirow}

\title{Pruning Pretrained Encoders with a Multitask Objective}

\author{%
 Patrick Xia\thanks{Work done while an intern at Microsoft.} \\
  Johns Hopkins University\\
  \texttt{paxia@cs.jhu.edu} \\
  \And
  Richard Shin \\
  Microsoft Semantic Machines\\
  \texttt{richard.shin@microsoft.com} \\
}

\begin{document}

\maketitle

\begin{abstract}
 The sizes of pretrained language models make them challenging and expensive to use when there are multiple desired downstream tasks. In this work, we adopt recent strategies for model pruning during finetuning to explore the question of whether it is possible to prune a single encoder so that it can be used for multiple tasks. We allocate a fixed parameter budget and compare pruning a single model with a multitask objective against the best ensemble of single-task models. We find that under two pruning strategies (element-wise and rank pruning), the approach with the multitask objective outperforms training models separately when averaged across all tasks, and it is competitive on each individual one. Additional analysis finds that using a multitask objective during pruning can also be an effective method for reducing model sizes for low-resource tasks.
 
\end{abstract}

\section{Introduction}

NLP models typically use pretrained text encoders like BERT \citep{devlin-etal-2019-bert}, which perform well when finetuned across many downstream NLP tasks. At the same time, these models are often overparameterized for the downstream task, leading to a surge of interest in reducing encoder size while retaining most of its performance on downstream tasks \citep{sun-etal-2020-mobilebert, sanh2020distilbert}. 

Meanwhile, there has been interest in adapting a \textit{single} model to \textit{multiple} downstream tasks through the use of a small number of additional, 
task-specific parameters \citep{pmlr-v97-houlsby19a, shin-etal-2020-autoprompt, Hu2021LoRALA}. These techniques are useful for efficiently sharing large base models during training by freezing the underlying encoder and finetuning parameters dependent on the target task. Switching between tasks is also cheap: only a small component is changed. These methods for extending models of size $N$ with $\epsilon$ parameters per task can perform well on $t$ tasks at the cost of $N + t\epsilon$ instead of $tN$ parameters, and architectural innovations have led to smaller $\epsilon$ \citep{karimi-mahabadi-etal-2021-parameter, mahabadi2021compacter}. In practice, this allows for deployment of models for multiple tasks at the cost of a single model in terms of memory or disk space. 

In this work, we explore the possibility of further reducing the number of parameters used by these multitask models to be substantially smaller than $N$. Specifically, we aim to do so by pruning  \textit{multitask} models using a multitask training objective.  In addition to the benefits described above, we aim for the best of both worlds: a substantially pruned model that also performs well on multiple tasks. We also ask whether individual task performances can be improved by leveraging data from the other tasks, which is a strategy employed by general-purpose language modeling \citep{liu-etal-2019-multi,aghajanyan2021muppet}.


We present an empirical study specifically for pruning in the multitask scenario. 
In this paper:

\begin{itemize}
    \item We extend both structured and unstructured pruning methods to the multitask setting.
    \item Under both methods, we find that a multitask model consistently outperforms a combination of single-task models for a given budget.
    \item Using a multitask objective does not necessarily lead to a loss in performance on any individual task, and in some cases, can improve upon a single-task objective.
    \item A multitask objective enables improved performance for tasks with smaller dataset sizes.
\end{itemize}

\section{Approach}

\subsection{Pruning Methods}
\label{sec:model_pruning}

We explore 8 settings for model pruning, varying pruning method (magnitude vs. movement), what gets pruned (element-wise (unstructured) pruning vs. rank (structured) pruning), and where (global vs. local). \autoref{tab:methods} summarizes the differences. We use the term \textit{sparsity} to refer to the fraction of removed (or zeroed out) weights.



\paragraph{Magnitude vs. movement pruning} We consider two pruning heuristics.
\textit{Magnitude pruning} is a well-studied method which incrementally removes (sets to zero) the smallest-magnitude weights during training, until only the top-$k$ largest weights remain at completion \citep{LeCun1989OptimalBD, 10.5555/2969239.2969366, h.2018to}. In this work, $k$ refers to a \textit{fraction} of the weights, and so it directly controls the sparsity level and size of the pruned model.

\textit{Movement pruning} \citep{DBLP:conf/nips/Sanh0R20} is a first-order method which prunes by using learned importance scores that correspond to the direction of growth of the weights. Formally, for a weight $\mathbf{W}$, this method associates with it a set of scores $\mathbf{S}$. In the forward pass, a binary mask $\mathbf{M} = \text{Top}_k(\mathbf{S})$ is applied to the weight, and so for an input $\mathbf{x}$, $(\mathbf{W} \odot \mathbf{M})\mathbf{x}$ is used instead, where $\odot$ is the element-wise product. In the backward pass, all weights are updated as $\text{Top}_k$ is ignored and its gradient is approximated using a straight-through estimator \citep{Bengio2013EstimatingOP}. \citet{DBLP:conf/nips/Sanh0R20} showed this to be effective during finetuning when the model's pretraining objective is different from that of a downstream task. 

There are extensions to both methods which use an minimum threshold rather than top-$k$; however, controlling the sparsity then requires a sweep per task to discover the best threshold, which is expensive especially in the multitask setting.

\paragraph{Unstructured vs. structured pruning} 
In prior work, weights are often pruned per-parameter, or \textit{element-wise} based on a heuristic like top-$k$. This can lead to arbitrary, unstructured, sparsity patterns. It can be difficult to realize practical efficiency gains from doing so as arbitrary locations in weight matrices will be set to zero.
Storing such sparse matrices does not lead to immediate gains and sparse matrix multiplication is not always faster, especially on GPUs \citep{gale10.5555/3433701.3433723}.
As such, other work considers structured pruning of entire rows or columns of the matrices, which makes it much easier to realize efficiency gains \citep{fan2021layerwise, lagunas2021block}.
We explore an  alternative structured pruning approach, \textit{rank pruning} \citep{Yang_2020_CVPR_Workshops}. 

Starting with a weight $\mathbf{W} \in \mathbb{R}^{m \times n}$, we use SVD to approximate $\mathbf{W} = \mathbf{U}\mathbf{\Sigma}\mathbf{V}$ where $\mathbf{U} \in \mathbb{R}^{m \times k}$, $\mathbf{\Sigma} \in \mathbb{R}^{k \times k}$ is diagonal, and $\mathbf{V} \in \mathbb{R}^{k \times n}$. Initially, $k = \min(m, n)$. By pruning values from the diagonal of $\mathbf{\Sigma}$, we reduce its rank from $k$ to $k'$. Consequently, entire columns in $\mathbf{U}$ and rows in $\mathbf{V}$ can be pruned, resulting in three reduced matrices: $\mathbf{U'}$, $\mathbf{\Sigma'}$, and $\mathbf{V'}$. 

The resulting matrices can be stored as $\mathbf{U'}\mathbf{\Sigma'} \in \mathbb{R}^{m \times k'}$ and $\mathbf{V'} \in \mathbb{R}^{k' \times n}$, which uses $k'(m+n)$ parameters. If $k' < \frac{k}{2}$, this is smaller than the original $m \times n$ size of $\mathbf{W}$.\footnote{
If $k' > \frac{k}{2}$, we can recover the unfactorized form, $\mathbf{W'} = \mathbf{U'}\mathbf{\Sigma'} \mathbf{V'}$. Thus, rank pruning would never use more parameters or computation than element-wise pruning.} Similarly, decomposing $\mathbf{W}$ and pruning in this way would lead to faster inference -- for an input $x \in \mathbb{R}^{m \times l}$, $\mathbf{W}^\top x$ costs $O(lmn)$ while $((\mathbf{U'}\mathbf{\Sigma'})\mathbf{V'})^\top x$ only needs $O(lk'(m + n))$.  

The elements in $\mathbf{\Sigma}$ are initially the singular values of $\mathbf{W}$, and so their magnitudes are interpretable as the importance of a particular dimension. \citet{Yang_2020_CVPR_Workshops} prune directly on the magnitudes while enforcing orthogonal regularizers to constrain $\mathbf{\Sigma}$ to be close to the singular values. In contrast, we only use SVD as initialization and use no additional regularizers. Since it is not obvious that interpreting the weights as singular values is valid after finetuning, we also try movement pruning as it is empirically effective for finetuning.

\paragraph{Global vs. local pruning} We experiment with pruning both \textit{locally} (each weight is pruned to a target sparsity) and \textit{globally} (the entire model is pruned to a target sparsity). Pruning globally allows for more aggressive pruning of less useful model components or layers. However, it could also be harmful if the weights are not globally comparable. In particular, each Transformer layer has self-attention and feedforward weights that are not normalized to the same scale, especially after SVD.


\begin{table*}
    \centering
    \small
    \begin{tabular}{lll}
    \toprule
    Variable & Choices & How it changes pruning objective \\
    \midrule
    \multirow{2}*{Selection} & Magnitude & Remove model weights with smallest magnitudes \\
    & Movement & Learned importance scores per weight \\
    \midrule
    \multirow{2}*{Structure} & Element-wise &  Prune parameters independently from each other\\
    & Rank & SVD then prune diagonal entries (rank) \\
    \midrule
    \multirow{2}*{Scope} & Global &  top-$k$ computed across entire model \\
    & Local & top-$k$ computed per weight matrix \\
    \bottomrule
    \end{tabular}
    \caption{Summary of the 8 settings for model pruning we explored. There are $3$ variables with $2$ choices each, leading to $2^3 = 8$ combinations.
    In all settings, the pruning methods all iteratively prune the model to keep a constant fraction of parameters.
    }
    \label{tab:methods}
\end{table*}

\subsection{Multitask extension}

Extending these methods to the multitask setting is straightforward. We train separate, unpruned classification heads for each task while sharing a common set of pruned encoder weights (and learned importance scores, in the case of movement pruning) across all tasks.  

Alternatively, one could selectively prune or modify weights (and learn importance scores) based on the specific task \citep{liang-etal-2021-super, guo-etal-2021-parameter}. In preliminary experiments, we explore this idea by considering three methods of creating the task-specific mask: 1) \textit{shared}, in which $\mathbf{M}_{t_i} = \mathbf{M}_{t_j}$ for all tasks $t_i, t_j$, and so there would only be one set of scores, $\mathbf{S}$; 2) \textit{separate}, in which for a task $t$, a separate scores $\mathbf{S}_t$ is learned and updated, which can then produce a corresponding $\mathbf{M}_t$; 3) \textit{hybrid}, in which both a shared $\mathbf{S}$ and a task-specific $\mathbf{S}_t$ are learned and the corresponding mask $\mathbf{M}_t = \max(\text{Top}_k(\mathbf{S}), \text{Top}_k(\mathbf{S}_t)).$ 

However, we found both \textit{separate} and \textit{hybrid} to be expensive or slow during training when there are too many tasks, as a separate $S_t$ needs to be stored, possibly outside of limited GPU memory. At the same time, we found that they did not lead to performance improvement. Especially under \textit{separate}, there was minimal sharing of parameters between tasks. Furthermore, this idea cannot be easily extended to magnitude pruning. Therefore, for the remainder of the study, we focus on the simplest \textit{shared} setting.


In multitask pruning, each task is sampled uniformly at random and optimized for that task's objective. While there may be some benefits to smoothing the sampling based on the dataset size (e.g. see Section 3.5.2 of \citet{JMLR:v21:20-074}), we focus on the simplest setting for this proof-of-concept study.

\subsection{Model and Datasets}

Like prior work \citep{jiao-etal-2020-tinybert, sun-etal-2020-mobilebert, DBLP:conf/nips/Sanh0R20}, we prune the BERT-base model \citep{devlin-etal-2019-bert} and evaluate on a variety of downstream NLP benchmarks. 
We report sparsity relative only to the 12-layer Transformer, and we consider the embedding and output classification layers a fixed cost. This is also the treatment in \citet{DBLP:conf/nips/Sanh0R20}. We use their code with the same hyperparameters for all main experiments, with the exception of rank pruning for which we increase the learning rate of $\mathbf{\Sigma}$ to $5 \times 10^{-3}$. Like that work, we also primarily benchmark on MNLI \citep{williams-etal-2018-broad}, SQuADv1.1 \citep{rajpurkar-etal-2016-squad}, and QQP \citep{WinNT}.

\section{Experiments and Results}

First, we benchmark each of the 8 pruning methods on a single dataset  (Section \ref{sec:experiments:comp}). The goal is to identify task and practical tradeoffs between structured and unstructured pruning and the best settings for each. In Section \ref{sec:experiments:multi}, we compare a model pruned with a multitask objective against an ensemble of single-task models to explore whether a multitask objective can be effective. Finally, we ask whether a multitask objective can provide benefits as an \textit{auxiliary} objective even in the single-task setting (Section \ref{sec:experiments:single_task}).

\subsection{Comparing Pruning Strategies}
\label{sec:experiments:comp}

\begin{figure*}
    \centering\subfloat{\label{fig:mnli}\includegraphics[width=100mm]{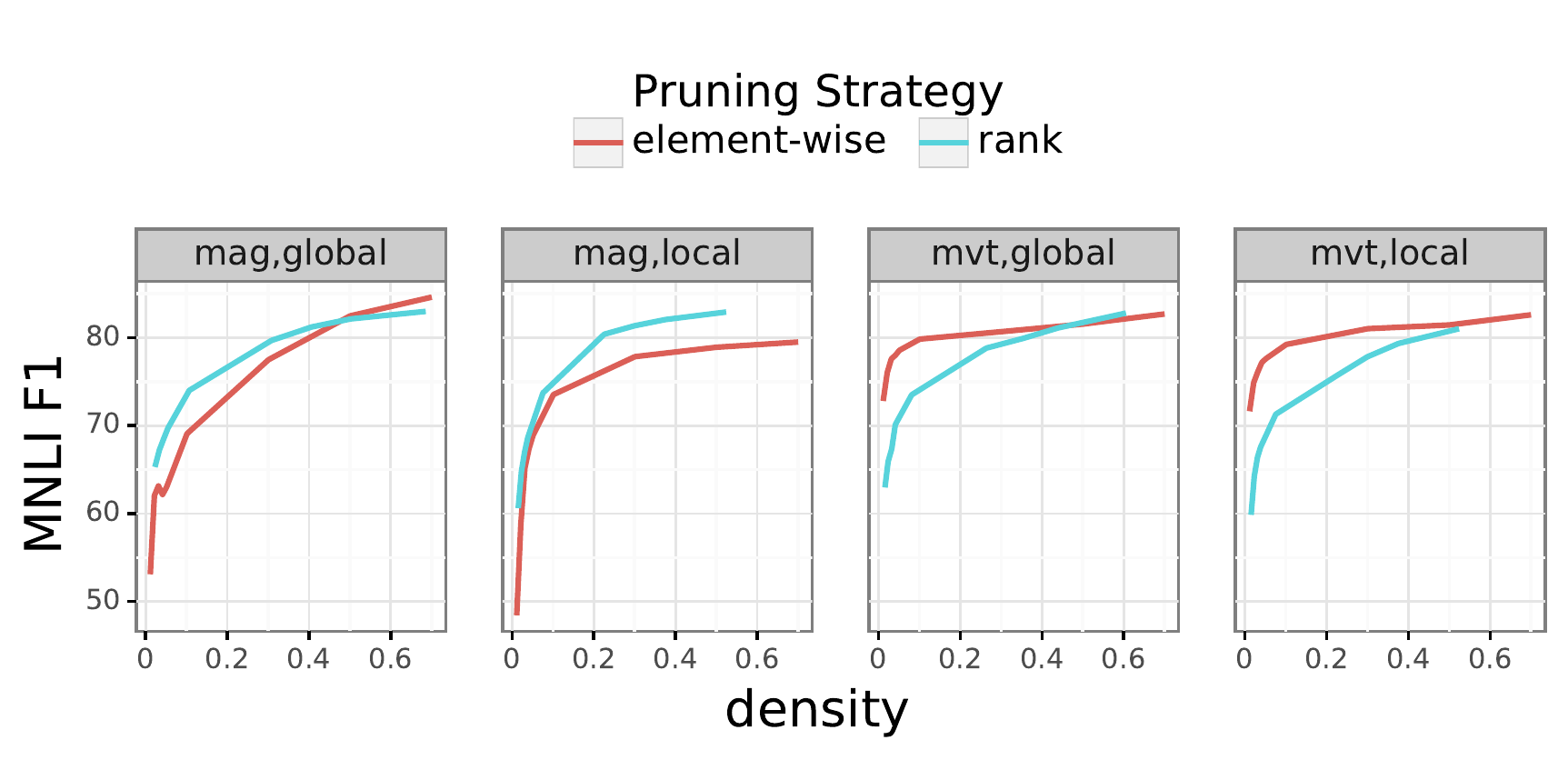}}
\subfloat{\label{fig:speed}\includegraphics[width=30mm]{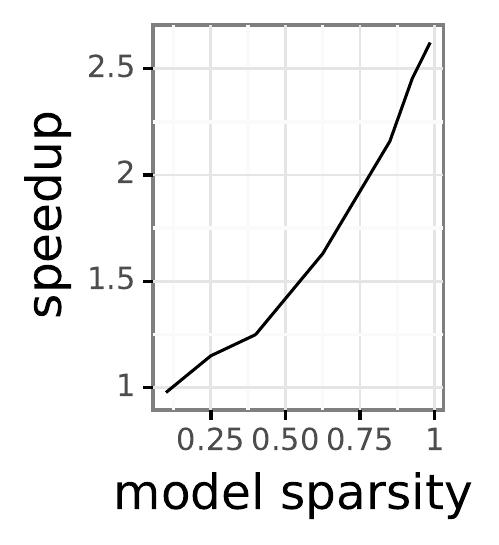}}
    \caption{Left: The performance on MNLI (dev) across [magnitude, movement] and [global, local] pruning strategies. Within each plot, we show the performance of pruned element-wise and rank-pruned models. Right: Comparison of the runtime of a model using rank pruning relative to an entirely dense model, showing that the structured sparsity ensured by rank pruning can lead to practical benefits. \textit{Density} is 1 - sparsity.}
    \label{fig:runtime}
\end{figure*}

We establish the relative task performance difference between element-wise and rank pruning. We prune BERT-base to different sparsity levels while finetuning on MNLI for each of the parameter combinations described in Section \ref{sec:model_pruning}. We perform this in the single-task setting to compare the possible benefits that can be derived from rank pruning and to find the best configuration for both methods. Figure \ref{fig:mnli} presents both task performance (averaged F1) and inference speed.\footnote{The runtime comparison was performed on a single NVIDIA RTX 6000.}

\paragraph{Task performance} 
These results in element-wise pruning once again corroborate the superiority of global movement pruning found by \citet{DBLP:conf/nips/Sanh0R20}. In contrast, we find that for rank pruning, local magnitude pruning is the best strategy at high sparsity levels. The poor performance of global pruning can be explained by the fact that it is possible to prune some parameters marginally (i.e. not beyond the $\frac{k}{2}$ threshold needed to see improvements) and so some of the budget allocated towards pruning is unused. Furthermore, the magnitudes of the singular values of the feedforward and self-attention layers are not necessarily comparable, as we find that attention weights are pruned more aggressively. 


Comparing between the best settings for unstructured (element-wise) and structured (rank) pruning, we find that unstructured pruning retains performance better when the parameter count is low. 

\paragraph{Runtime comparison} 
To compare the runtime of element-wise and rank pruning, we make a simplifying assumption that unstructured pruning does not shrink the size of any individual matrix, and so the runtime at any sparsity level would be equal to that of an unpruned model. Under this assumption, we find in Figure \ref{fig:speed} that rank pruning offers substantial speedups and is almost immediately (at around 10\% sparsity) faster than the dense model. 

However, sparsifying a matrix can lead to specialized hardware and algorithmic optimizations as demonstrated by sparse multiplication libraries \citep{gale10.5555/3433701.3433723}. \citet{lagunas2021block} optimize element-wise unstructured pruning in a simple manner by removing entirely pruned rows, columns or attention heads. They show that even at high sparsities (more than 90\%), this strategy achieves at most around a 1.5$\times$ speedup. Meanwhile, rank pruning achieves a comparable speedup at just 50\% and almost a 2.5$\times$ speedup at 90\%. While rank pruning has lower F1 under a fixed parameter budget, it is a competitive option given a fixed \textit{latency} budget. 


\paragraph{Global rank pruning} One limitation of global rank pruning is that the initialization is not normalized across weights. Specifically, we observe that with \textit{magnitude} as our selection heuristic, attention weights are pruned more aggressively because most of their singular values are smaller than those of the feedforward parameters. This could result in important weights in the attention layers being pruned before less important ones in the feedforward layers, and it could explain why global rank pruning performs poorly at high sparsities (as shown in Figure \ref{fig:mnli}). The opposite behavior is observed for rank \textit{movement} pruning, where pruning globally is slightly preferred over pruning locally. This suggests that first-order information might be more comparable globally. 

Future work can investigate finding a balance between using the interpretability of the magnitudes of the singular values and first-order information from the finetuning process. Alternatively, one could explore a hybrid pruning strategy: with \textit{magnitude} pruning, separate thresholds (or $k$) for the attention and the feedforward parameters. 



\begin{figure*}
\centering     
\subfloat[Unstructured (element-wise) pruning]{\label{fig:unstructured}\includegraphics[width=70mm]{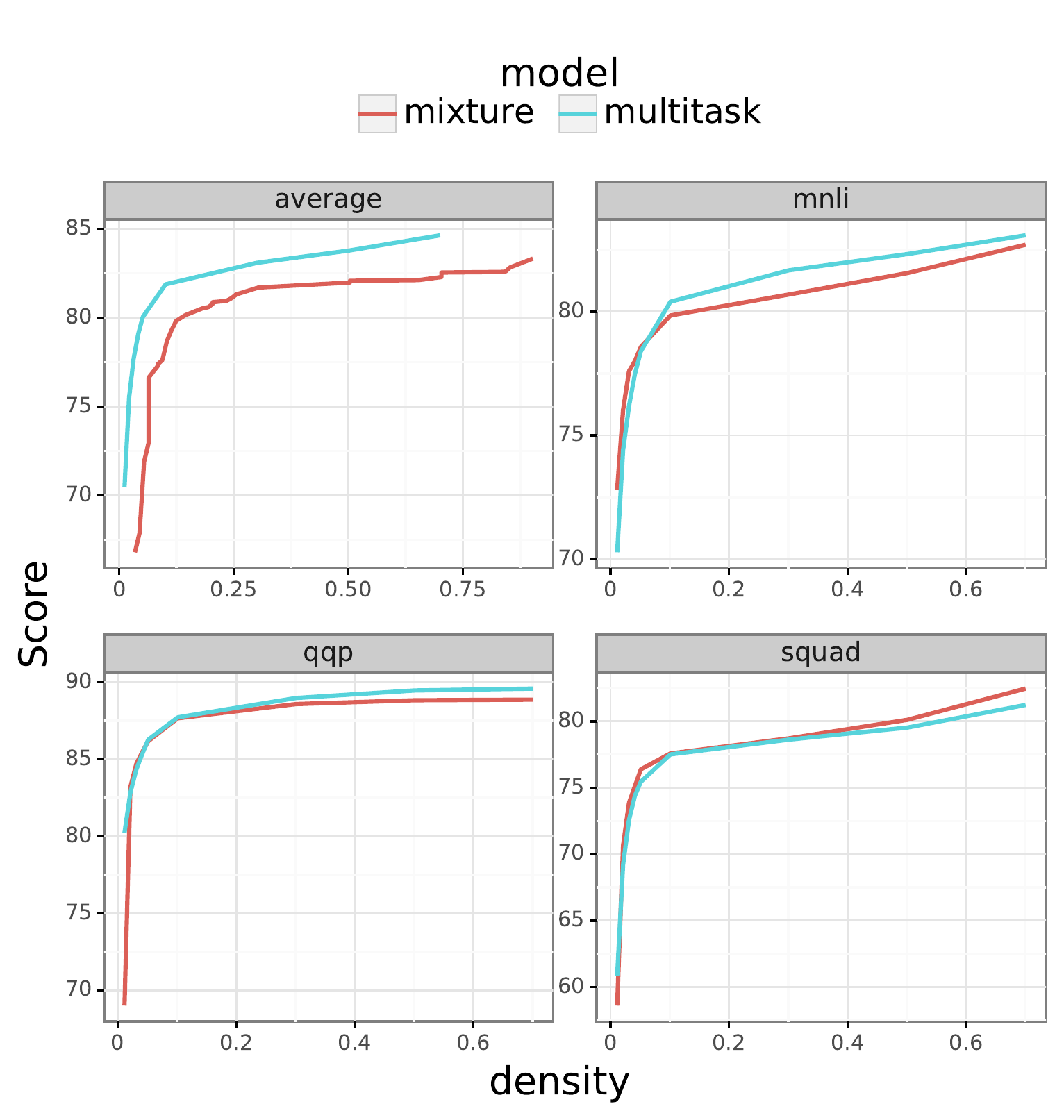}}
\subfloat[Structured (rank) pruning]{\label{fig:structured}\includegraphics[width=70mm]{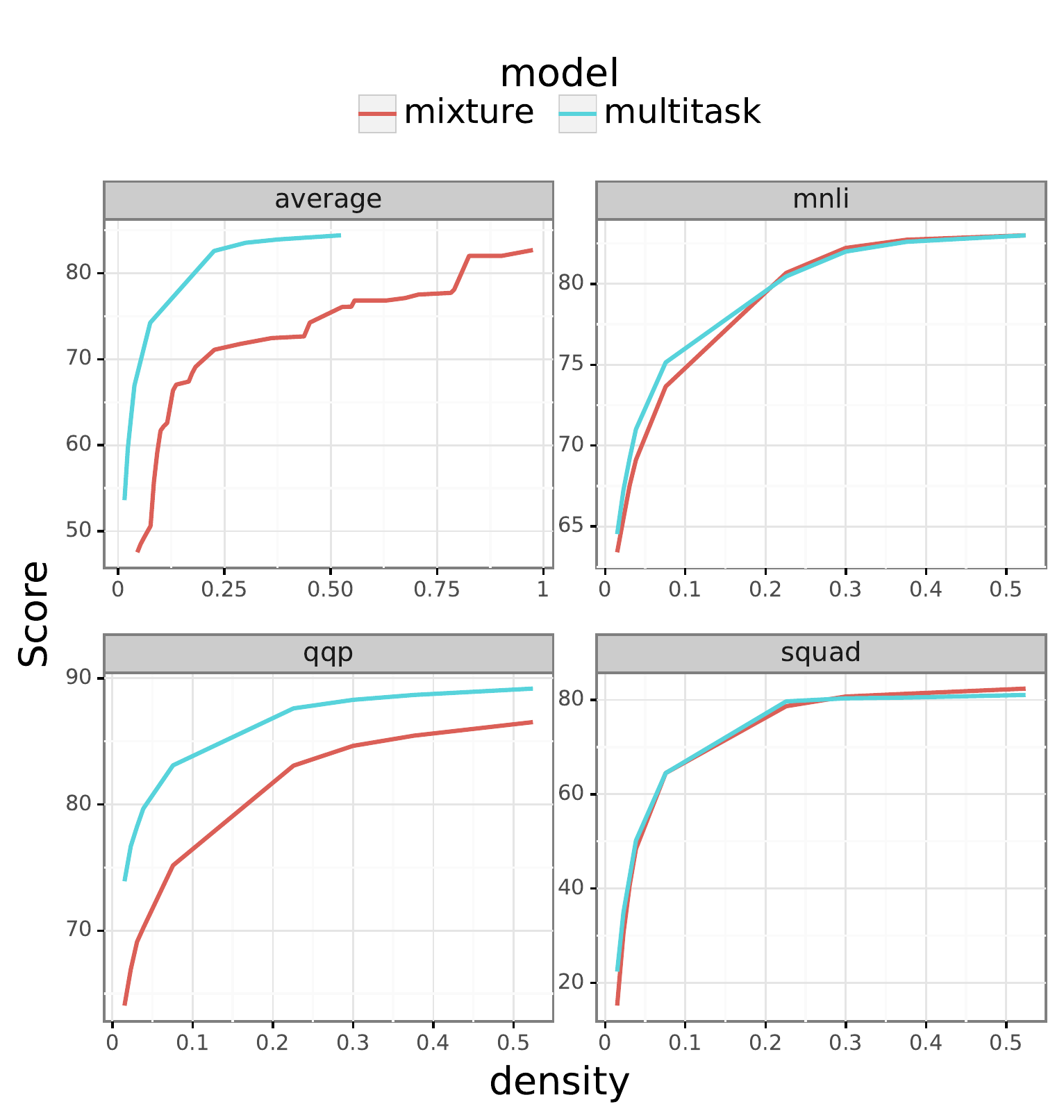}}
\caption{Given a fixed parameter budget (expressed as a fraction of a single model size), we compare a single model pruned with a multitask objective (blue) and the best combination of 3 individual task models for a given size. The red line (``mixture'') is the Pareto frontier of these combinations.}
\label{fig:multitask}
\end{figure*}

\subsection{Multitask pruning}
\label{sec:experiments:multi}

With the above baseline established, we conduct experiments in a multitask setting with MNLI, QQP, and SQuAD. Our goal is to answer: given a fixed parameter budget, is it better to train (and prune) three specialized models or one multitask model for all three tasks? Note that training three specialized (single-task) models means that each model, on average, would need to be a third of the size of one multitask model.

Concretely, we prune models to different sparsity levels using four separate objectives: MNLI, QQP, SQuAD, and multitask. We compare the best mixture of single-task models to a single model pruned with a multitask objective.\footnote{We train for 8 epochs with 2 initial and 2 final warmup epochs. This is close the average of the hyperparameters previously found for the three separate tasks.} The best mixture is determined by the Pareto frontier of all possible ensembles of models. \autoref{fig:multitask} shows that the multitask model outperforms the mixture on a macro-averaged 3-task metric. In addition, it matches or exceeds the performance on individual tasks. 

We find that multitask pruning outperforms the mixture with both element-wise pruning and rank pruning, suggesting that the ability to leverage a multitask objective during pruning may extend to other, novel methods for pruning. 
This also provides evidence that in encoders, there exist smaller shared subnetworks that can be leveraged across multiple tasks.

\begin{table*}[t]
    \centering
    \small
    \resizebox{\linewidth}{!}{%
    \begin{tabular}{@{}lllcccccccccc@{}}
    \toprule
        \bf{Model} & \bf{Rank} & \bf{Prune} \% & \bf{MNLI} & \bf{SQuAD} & \bf{QQP} &  \bf{QNLI} & \bf{SST-2} & \bf{MRPC} & \bf{STSB} & \bf{RTE} & \bf{CoLA} & \bf{Avg.} \\
        & & & 393K & 87.6K & 364K & 105K & 67K & 3.7K & 7K & 2.5K & 8.5K \\
    \midrule
    RP-3 & 38 & 7.6\% & 73.8 & 64.3 & 83.1 & - & - & - & - & - & - & - \\
    RP-1 & 38 & 7.6\% & 73.6 & 64.4 & 75.2 & 80.0 & 87.4 & 75.7 & 14.4 & 52.7 & 0 &  58.2 \\
    BNG & 150 & 29.2\% & - & - & - & - & 91.3 $\pm$ 0.4 & 87.8 $\pm$ 0.6 &  - & - & 38.7 $\pm$ 1.6 & - \\
    \addlinespace
    RP-9 & 38 & 7.6\% & 72.1 & 54.2 & 81.3 & 84.7 & 87.6 & 83.6 & 86.3 & 67.1 & 23.6 & 71.2 \\
    RP-9 & 76 & 15.0\% & 76.8 & 70.5 & 84.5 & 88.2 & 89.8 & 87.5 & 88.0 & 70.4 & 34.6 & 76.7\\
    \bottomrule\\
    \end{tabular}%
    }
    \caption{Individual task performance (dev.) of a model pruned using rank pruning with the multitask objective. The size of the training set for each task is also listed. RP-9 is trained on all nine tasks, RP-3 is the three-task model from Section \ref{sec:experiments:multi}, and RP-1 represents 9 separate single-task baseline models. BNG \citep{ben-noach-goldberg-2020-compressing} is three separate single-task low-rank models tuned using knowledge distillation.}
    \label{tab:glue}
\end{table*}

\subsection{Multitask training as an auxiliary pruning objective}
\label{sec:experiments:single_task}

In Section \ref{sec:experiments:multi}, we find that in some cases, a multitask pruning objective is helpful even when only one task (e.g. QQP) is of interest.  
This finding, along with prior work demonstrating the effectiveness of multitask training \citep{wang-etal-2019-tell, liu-etal-2019-multi, aghajanyan2021muppet}, motivates our next question: if we only care about a single task, should we still use a multitask objective?

To test this hypothesis, we expand the three tasks to 9 and prune a model with a 9-task objective, sampling each task uniformly at random. 
In addition to the three aforementioned tasks, we include CoLA \citep{warstadt-etal-2019-neural}, SST-2 \citep{socher-etal-2013-recursive}, MRPC \citep{dolan2005automatically}, STSB \citep{cer-etal-2017-semeval}, and QNLI \citep{rajpurkar-etal-2016-squad}, and RTE \citep{dagan2005pascal}. For comparison, we also prune a model for each task separately. These additional models are pruned for 10 epochs, keeping other hyperparameters the same as the multitask model.

\autoref{tab:glue} shows the performance on each of the tasks when using local magnitude rank pruning. 
We find that the multitask models (RP-9) perform well on the smaller datasets of RTE, CoLA, and STSB, outperforming the single-task baseline (RP-1) and coming close larger models from prior work which relied on a hyperparameter search for each task \citep{ben-noach-goldberg-2020-compressing}. In contrast, we perform no hyperparameter tuning (beyond finding the best combinations in Section \ref{sec:experiments:comp}). These results collectively suggest that multitask-based pruning offers a way to effectively prune models for low-resourced tasks without extensive hyperparameter search.


Complementary to our approach of using multitask training is a set of methods in which a small number of additional task-specific parameters are directly embedded in the model, such as prefix embeddings \citep{keskarCTRL2019, li-liang-2021-prefix}, adapters \citep{pmlr-v97-houlsby19a}, diff parameters \citep{guo-etal-2021-parameter}, and LoRa layers \citep{Hu2021LoRALA}, among others. These parameters and specialized toward a specific task and can cheaply provide additional benefits. Future work can explore incorporating these task-specific modules into (or after) the pruning process.


\section{Conclusion}
\label{sec:conclusion}

We explore multitask pruning, which aims to perform well on multiple tasks with a given parameter budget smaller than the size of single-task model. Using methods for both unstructured (element-wise) pruning and structured (rank) pruning strategies, we find that pruning with a multitask objective outperforms combining multiple models pruned separately on individual task objectives. Additionally, we find that pruning with a multitask objective (without additional hyperparameter tuning) can help further prune tasks with low-resource datasets. We hope that these lessons learned guide future work in compressing multitask models.

\section*{Acknowledgements}
We would like to thank Jason Eisner, Hao Fang, Edward Hu, Anthony Platanios, Yu Su, Sam Thomson, and Benjamin Van Durme for helpful discussions and feedback for this project.

\bibliographystyle{acl_natbib}
\bibliography{anthology,custom}

\end{document}